\documentclass[final]{cvpr}

\usepackage{times}
\usepackage{epsfig}
\usepackage{graphicx}
\usepackage{amsmath}
\usepackage{amssymb}
\usepackage{helvet}   
\usepackage{courier}  
\usepackage{url}      
\usepackage{xcolor}
\usepackage{pifont}
\usepackage{booktabs}
\usepackage{multirow}
\usepackage{algorithmic}
\usepackage[linesnumbered,ruled,vlined]{algorithm2e}
\usepackage[export]{adjustbox}
\usepackage{color}
\usepackage{tabularx}
\usepackage[super]{nth}

\DeclareMathSymbol{\mh}{\mathord}{operators}{`\-}

\usepackage[pagebackref=true,breaklinks=true,colorlinks,bookmarks=false]{hyperref}

\begin{document}

\title{Multi-Modality Task Cascade for 3D Object Detection}

\author{Jinhyung Park,~ Xinshuo Weng,~ Yunze Man,~ Kris Kitani \\
Carnegie Mellon University \\
{\tt\small jinhyun1@andrew.cmu.edu~~\{xinshuow, yman, kkitani\}@cs.cmu.edu}
}
\maketitle

\begin{abstract}
Point clouds and RGB images are naturally complementary modalities for 3D visual understanding - the former provides sparse but accurate locations of points on objects, while the latter contains dense color and texture information. Despite this potential for close sensor fusion, many methods train two models in isolation and use simple feature concatenation to represent 3D sensor data.
This separated training scheme results in potentially sub-optimal performance and prevents 3D tasks from being used to benefit 2D tasks that are often useful on their own.
To provide a more integrated approach, we propose a novel Multi-Modality Task Cascade network (MTC-RCNN) that leverages 3D box proposals to improve 2D segmentation predictions, which are then used to further refine the 3D boxes. We show that including a 2D network between two stages of 3D modules significantly improves both 2D and 3D task performance. 
Moreover, to prevent the 3D module from over-relying on the overfitted 2D predictions, we propose a dual-head 2D segmentation training and inference scheme, allowing the second 3D module to learn to interpret imperfect 2D segmentation predictions. Evaluating our model on the challenging SUN RGB-D dataset, we improve upon state-of-the-art results of both single modality and fusion networks by a large margin (\textbf{+3.8} mAP@0.5). Code will be released \href{https://github.com/Divadi/MTC_RCNN}{here.}
\end{abstract}

\section{Introduction}
\begin{figure}[t]
  \centering
  \includegraphics[width=\linewidth]{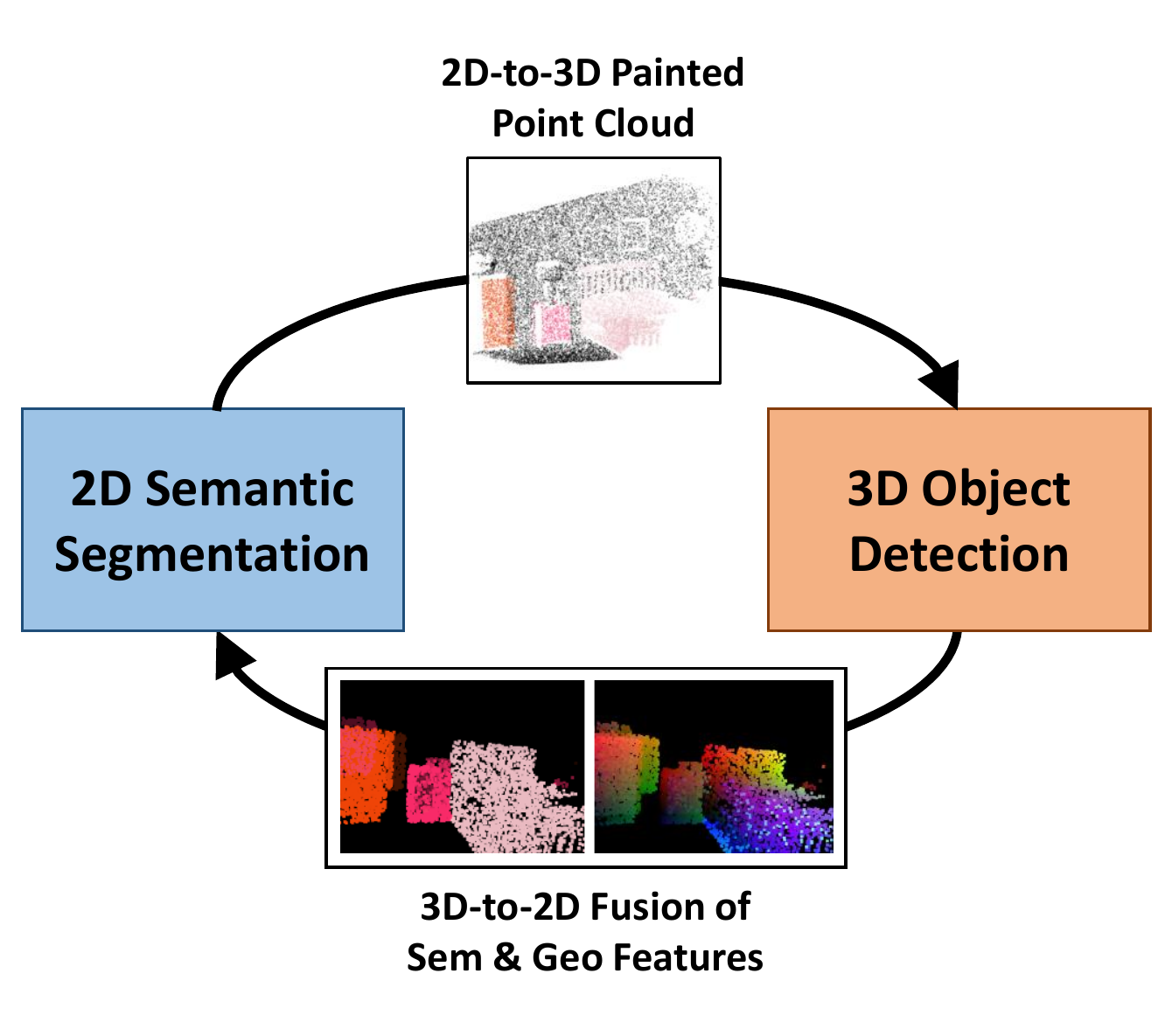}
  \caption{\textbf{Recursive cascade of 2D semantic segmentation and 3D object detection.} Rich semantic features from 2D segmentation on dense 2D images can benefit 3D object detection on sparse 3D point clouds. Then, semantic and structural information from these strengthened 3D box predictions can further improve 2D segmentation. Again conversely, this improved 2D segmentation can refine the original 3D box predictions. This recursive pipeline can be repeatedly applied.}
  \label{fig:teaser}
\end{figure}

3D detection requires precise localization of objects in three-dimensional space, which is made difficult by the inherent sparsity and noise present in point clouds when using LiDAR as the sole input source. Many objects have only a few points captured on them due to either distance, occlusion, or reflectivity, causing structurally similar objects to be indistinguishable in the point cloud. On the other hand, RGB images have a far higher resolution than point clouds, containing a dense array of color and texture cues. Objects with only a few points in 3D can occupy a magnitude more pixels in 2D, allowing richer semantic features to be extracted for these objects than can be extracted from 3D. Conversely, features with 3D structural information obtained from spatially reasoning on point clouds can in turn complement RGB images as well, which lack explicit depth information. These observations support our intuition that point clouds and RGB images are \textit{mutually} beneficial modalities.

Recent work proposes different methods of fusing image and point cloud semantics for 3D detection. Some works use a pre-trained 2D detector to generate initial frustum-based region proposals \cite{Qi2018FrustumPF, Wang2019FrustumCS}. 
Alternatively, instead of constraining the 3D search space, other methods propose to use 2D features to enrich point features. Some methods fuse semantics obtained at the end of a 2D network, either using 2D task predictions \cite{Qi2020ImVoteNetB3, Xie2020PIRCNNAE, Vora2020PointPaintingSF, Pang2020CLOCsCO} or using 2D backbone features \cite{Sindagi2019MVXNetMV, Yoo20203DCVFGJ}. Another line of work \cite{Liang2018DeepCF, Huang2020EPNetEP} fuse features from every layer of the 2D network. Despite demonstrating improvements over 3D-only methods, the image and point cloud fusion methods for 3D detection in prior work exploit only half of the mutually beneficial relationship between these two modalities. These methods only extract 2D semantics to benefit 3D tasks and do not consider using 3D semantics for better 2D feature extraction. We argue that by additionally including 3D task predictions as input to a 2D image network, the resulting improved 2D semantics can better benefit the 3D task.

In this work, to leverage the cyclic, mutually complementary relationship between images and point clouds, we propose a new \textbf{M}ulti-modality \textbf{T}ask \textbf{C}ascade network for 3D object detection (\textbf{MTC-RCNN}). As illustrated in Figure \ref{fig:teaser}, the key idea of our approach is to include a 2D segmentation network \cite{Chen2018EncoderDecoderWA} \textit{between} the first and second stages of a 3D detection network allowing it to both \textit{benefit from} and \textit{improve} the 3D detection modules. This 2D network takes as input both the 2D RGB-D image as well as the 3D-to-2D projected point-level semantic and geometric features. Given a point, we obtain its features from the 3D proposal box it is in, using both the 3D box's class prediction and its box parameters. From our experiments, we observe that fusing these 3D features into the 2D network improves both 2D segmentation performance as well as the quality of the final 3D boxes. Further, the 2D network is supervised by ground truth generated from 3D box annotations, requiring no additional labels. 

After obtaining the 2D segmentation predictions, we refine the 3D proposal boxes via a simple second stage 3D network inspired by \cite{li2021lidar}. For each 3D proposal box, the points within it are projected onto the 2D segmentation predictions, and the corresponding channel-wise probability distribution is concatenated with the point's 3D box-based geometric features. A PointNet \cite{Qi2017PointNetDL} processes the resulting point features and refines the 3D proposal.

Different from some other multi-modality methods \cite{Qi2020ImVoteNetB3, Xie2020PIRCNNAE, Vora2020PointPaintingSF}, we jointly optimize the 2D and 3D networks, allowing them to learn better shared feature representations. However, we find that the 2D network tends to overfit faster than the 3D network, quickly converging to perfect segmentations. This causes the 2D segmentation predictions to dominate the 3D features in the \nth{2} stage 3D network during training. To alleviate this issue, we propose to instead fuse the segmentation predictions of a weaker, auxiliary head during training to allow the \nth{2} stage 3D network to learn to interpret imperfect 2D segmentation predictions. During testing, an ensemble of the auxiliary head and the DeepLabV3+ head \cite{Chen2018EncoderDecoderWA} is used.

Finally, we find that our framework is particularly well-suited for PointPainting \cite{Vora2020PointPaintingSF}. The addition of a 2D segmentation network before the first 3D network generates better proposals that not only directly benefit the 3D detection task but also improves 2D segmentation predictions which further improves 3D box refinement.

We evaluate our approach on the difficult SUN RGB-D dataset \cite{Song2015SUNRA}. Our models offer significant gains over our two-stage 3D-only baseline (+5.1 AP@0.25, +3.0 AP@0.50), validating the importance of adding a 2D segmentation network between the \nth{1} and \nth{2} stage 3D modules. We also provide ablations, investigating the difficulties of multi-modality training and justifying our design choices. MTC-RCNN improves upon state-of-the-art results of both single modality and fusion networks (\textbf{+1.2 AP@0.25}, \textbf{+3.8 AP@0.5}).

The contributions of this paper can be summarized as follows:
\begin{itemize}
    \itemsep -2pt
    \item Our novel image and point cloud fusion network fully leverages both directions of the mutually beneficial relationship between the two modalities.
    \item We investigate the difficulties of multi-modality training and propose a new direction of limiting the performance of one modality during training.
    \item Our method not only achieves new SOTA performance in 3D object detection on the SUN RGB-D dataset but also yields 2D segmentation predictions significantly better than a 2D-only baseline.
\end{itemize}

\label{sec:intro}

\section{Related Work}
\noindent\textbf{3D Object Detection on Point Clouds.}
To address the irregularity and sparsity of point clouds, one direction of research proposes to organize the points into either 2D or 3D grids to be further processed by CNNs. The early work MV3D \cite{Chen2017Multiview3O} processes multiple 2D projections and fuses them at a proposal level. VoxelNet \cite{Zhou2018VoxelNetEL} instead works with 3D voxels, using a 3D CNN to extract features. Followup methods \cite{Graham20183DSS, Yan2018SECONDSE, Choy20194DSC, Shi2020FromPT} exploit the sparsity of point clouds, only performing convolution operations on non-empty voxels. Following the works PointNet \cite{Qi2017PointNetDL} and PointNet++ \cite{Qi2017PointNetDH}, another line of research directly processes point clouds. PointRCNN \cite{Shi2019PointRCNN3O} generates 3D proposal boxes from predicted foreground points, while \cite{Qi2019DeepHV, Yang20203DSSDP3} use a "voting" mechanism to shift candidate points closer to object centers. To take advantage of both the efficiency of voxel-based methods and precision of point-based methods, some recent works use both \cite{Lang2019PointPillarsFE, Liu2019PointVoxelCF, Noh2021HVPRHV, Tang2020SearchingE3}. These 3D detection methods achieve great success \textit{without using RGB images}. To further improve performance over geometry-only methods, our work proposes a new method of fusing 2D images and point clouds.

\noindent\textbf{Point Cloud and Image Fusion-based 3D Object Detection.} Early methods seeking to leverage RGB images for 3D detection were 2D-driven, using mature 2D detectors to constrain the 3D search space before regressing 3D boxes \cite{Lahoud20172DDriven3O, Qi2018FrustumPF, Wang2019FrustumCS}. Observing that the performance of this paradigm is upper bounded by the 2D detector, other methods instead use image semantics to enhance 3D features. MV3D \cite{Chen2017Multiview3O} and AVOD \cite{Ku2018Joint3P} extract image features for 3D proposals, while \cite{Pang2020CLOCsCO} proposes to flexibly fuse 2D and 3D proposals. Methods can further be divided by whether they fuse features from multiple intermediate layers \cite{Huang2020EPNetEP, Liang2018DeepCF}, features from the end of a 2D model \cite{Chen2017Multiview3O, Ku2018Joint3P, Sindagi2019MVXNetMV, Yoo20203DCVFGJ}, or 2D task-level predictions \cite{Qi2020ImVoteNetB3, Xie2020PIRCNNAE, Vora2020PointPaintingSF}. UberATG-MMF \cite{Liang2019MultiTaskMF} uses all of the above. Our method is most closely related to works that fuse 2D task-level predictions, but we differ from prior methods in critical areas. First, most methods choose to freeze the 2D model \cite{Vora2020PointPaintingSF, Qi2018FrustumPF, Qi2020ImVoteNetB3, Xie2020PIRCNNAE} when training the 3D detector, but we train end-to-end. Second, our 2D network takes 3D proposals as input to improve the 2D task predictions to further improve 3D box refinement.

\noindent\textbf{Multi-Modality Fusion Training.} Outside of detection, multi-modal fusion has been investigated in domains of visual question answering \cite{Agrawal2015VQAVQ}, action recognition \cite{Kay2017TheKH}, and acoustic event detection \cite{Gemmeke2017AudioSA}. A recent work \cite{Wang2020WhatMT} analyzes some of the difficulties of multi-modal training, finding that naively incorporating multiple modalities can result in a drop in performance compared to single-modality networks, due to the fact that different modalities can hinder each other from properly training. We observe a similar but different challenge in our work, as \cite{Wang2020WhatMT} deals with late-stage concatenation of features of different modalities for a single task, while our method has a cascade of modalities for multiple tasks. In our work, we propose to resolve this issue by \textit{inhibiting} the performance of one modality on the train set so later modules in the cascade can learn to work with imperfect intermediate task predictions.
\label{sec:related}
\section{Method}

\begin{figure*}[t]
  \centering
  \includegraphics[width=\linewidth]{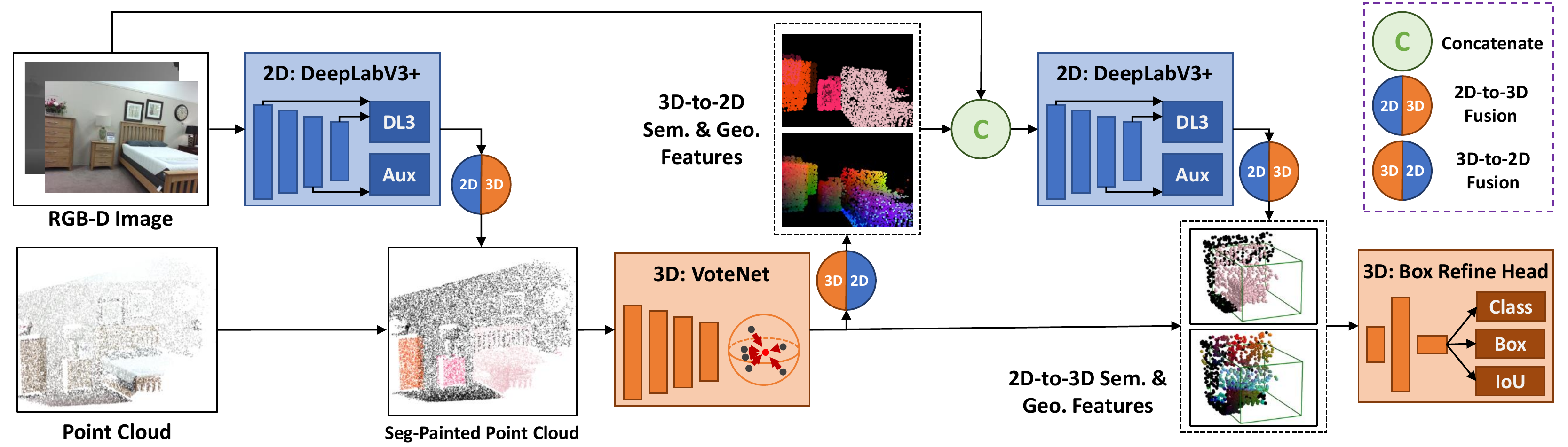}
  \caption{\textbf{Overview of our proposed MTC-RCNN.} The RGB-D image is processed by the \nth{1} 2D segmentation network, whose predictions are fused into the raw point cloud. Then, our \nth{1} 3D network, VoteNet, generates 3D proposals using this painted point cloud. Semantic \& geometric features are extracted from the 3D proposals, concatenated with the RGB-D image, and input into the \nth{2} 2D segmentation network. Finally, the \nth{2} 2D predictions are fused with the 3D proposals, which are refined by our \nth{2} 3D network.}
  \label{fig:main_arch}
\end{figure*}
\subsection{Overview\label{3.1}} 

MTC-RCNN is illustrated in Figure \ref{fig:main_arch}. At a high level, our multi-modality cascade can be described as (2D$\rightarrow$)3D$\rightarrow$2D$\rightarrow$3D, with the first 2D being the addition of PointPainting \cite{Vora2020PointPaintingSF}. In Sec. \ref{3.2}, we briefly outline VoteNet \cite{Qi2019DeepHV}, which we use for 3D proposal generation. In Sec. \ref{3.3}, we describe the 3D-to-2D fusion, the 2D segmentation architecture, and the method of generating 2D segmentation ground truth. Then, in Sec. \ref{3.4}, we present the method of 2D-to-3D fusion and the 3D box refinement network. Sec. \ref{3.5} describes how we apply PointPainting to our pipeline. Finally, Sec. \ref{3.6} outlines our training losses.

\subsection{3D Proposal Generation: Deep Hough Voting (VoteNet)\label{3.2}} 

For this work, we choose VoteNet to generate 3D proposals because it is a representative, 3D-only baseline. Given a set of points, VoteNet uses a PointNet++ \cite{Qi2017PointNetDH} backbone to extract features for a reduced set of sampled "seed" points. The seed features are then used to generate a set of "votes," with each vote consisting of a new 3D location closer to object centers as well as a new feature vector to be used for the final detection task. Finally, votes are clustered based on location, and each cluster predicts a 3D bounding box as well as its classification score. These 3D proposals are passed on to the later stages of our pipeline.

\subsection{3D-to-2D: Using 3D Proposals for 2D Semantic Segmentation\label{3.3}}

\noindent\textbf{3D-to-2D Fusion.} 
In order to use 3D task-level information to benefit the 2D segmentation task, given a set of 3D proposals, our method extracts point-level \textit{semantic} and \textit{geometric} features from them to include as input to the 2D semantic segmentation network. The semantic features help the 2D network reason about which class each point/pixel belongs to, while the geometric features helps encode information about objects' 3D structure. Given a 3D box proposal, we randomly sample a set of points contained within the box $\{p_i\}^n_{i=1}$, where each $p_i \in \mathbb{R}^3$ are 3D coordinates. For each point $p_i$, we obtain its semantic features $s_i \in \mathbb{R}^C$ by giving it the object class probability distribution of the 3D proposal, where $C$ is the number of classes. Then, to obtain geometric features, we transform each point to the box's canonical coordinates centered at the proposal's center and aligned to the proposal's heading direction. The geometric feature $g_i\in\mathbb{R}^9$ consists of two parts - the the point's canonical coordinates and the point's offset distances to the box's 6 surfaces. In all, for each proposal, we have a set of points sampled within the box as well as their semantic and geometric features:

\[
\{(p_i, f_i) \mid p_i \in \mathbb{R}^3, f_i = [s_i^\top, g_i^\top]^\top \in \mathbb{R}^{C+9}\}^n_{i=1} \tag{1}
\]
These points are then projected to the 2D image using the camera intrinsics, and their corresponding features are assigned to the projected 2D location, generating a 3D-to-2D feature map $\textbf{F}^{3D\mh to\mh 2D}$ of shape $H \times W \times (C + 9)$ ($H$ and $W$ are height and width of RGB-D image). We note that some points can be in multiple overlapping proposals, and we resolve this conflict by assigning each point to the highest confidence proposal it is in. Finally, $\textbf{F}^{3D\mh to\mh 2D}$ is concatenated channel-wise with the RGB-D image, resulting in a feature map $\textbf{F}^{2D\mh input}$ of shape $H \times W \times (C + 13)$ to be used as the input to our 2D segmentation network.

\noindent\textbf{2D Semantic Segmentation Network.} 
For our 2D network, we use DeepLabV3+ \cite{Chen2018EncoderDecoderWA} with a lightweight ResNet18 \cite{He2016DeepRL} backbone. We remove subsampling in the final two stages (C4 and C5) of the backbone and replace them with dilated convolutions. The DeepLabV3+ architecture includes a simple auxiliary head attached to the C4 stage, used for better supervision of intermediate layers. However, this head plays an important role in our work - we use predictions of this auxiliary head during training for 2D-to-3D fusion instead of the main head. We explain further in Sec. \ref{3.4} after we introduce our 3D proposal refinement module.

\noindent\textbf{2D Ground Truth Generation.} 
Here, we briefly describe the generation of 2D segmentation ground truth and provide more details in the supplementary. To generate 2D labels, we expand the 2D depth map to a 3D point cloud and assign points (pixels) within a 3D box the box's class label. We ignore pixels within multiple boxes and pixels that lack a depth value.

\subsection{2D-to-3D: Using 2D Segmentation for 3D Proposal Refinement\label{3.4}}

\textbf{3D Proposal Refinement Network.} 
From the first stage 3D network, we have a set of 3D proposal boxes. To refine these proposals, we extend a second-stage 3D refinement introduced by LiDAR-RCNN \cite{li2021lidar}. Given a proposal box, we enlarge it to capture more context. Then, we sample points within this enlarged box and extract each point's geometric features as described in Sec. \ref{3.3} - this yields a 9-dimensional feature vector for each point. 

In our 3D-only method, for each proposal, this set of geometric point features is run through a simple PointNet which predicts residuals for the box parameters as well the box's class type. Different from LiDAR-RCNN, this module is trained jointly, with gradients propagated through the per-point geometric features back to the first stage 3D network. Further, we add an IoU estimation branch to the PointNet to better the quality of the 3D box.

\noindent\textbf{2D-to-3D Fusion.} For 2D-to-3D fusion, we seek to use the 2D segmentation predictions from the 2D network as richer semantic features with which to guide the 3D box refinement. 
Extending the second stage 3D framework described above, for each proposal, the sampled points are projected onto the 2D segmentation predictions using the camera intrinsics. The corresponding channel-wise class distribution at the 2D projected point is appended to the 9-dimensional geometric features, resulting in each point having $9 + C$ dimensional features. These concatenated features are then passed through the PointNet as described previously.

During training, we fuse the 2D predictions of the auxiliary head because the 2D performance gap between training and testing is much smaller for the auxiliary head than it is for the main head. If we fuse the main head during training, the points around proposals are perfectly segmented, trivializing the refinement task. Fusing the auxiliary head during training and using both heads during testing, the model learns to adapt to imperfect 2D predictions.

\subsection{Early 2D-to-3D Fusion: PointPainting\label{3.5}} 

\begin{figure*}[t]
  \centering
  \includegraphics[width=\linewidth]{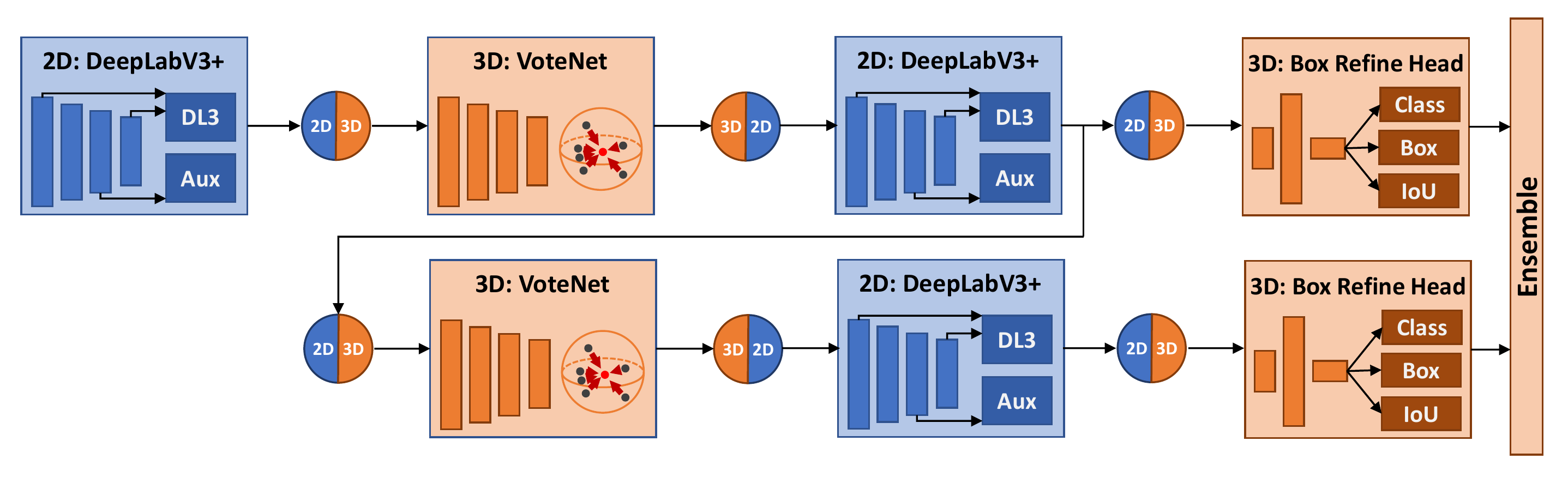}
  \caption{Illustration of recursively applying our pipeline twice.}
  \label{fig:recurse}
\end{figure*}
So far, we have presented a 3D-to-2D-to-3D pipeline consisting of the initial 3D proposal generation stage, the fusion 2D segmentation network, and the 3D refinement module. However, this cascade of modalities does not necessarily need to start with 3D detection. We can add a 2D segmentation network before the first 3D proposal generation following PointPainting \cite{Vora2020PointPaintingSF}. Specifically, before the point cloud is input into VoteNet, each point is projected onto initial 2D segmentation predictions and assigned the probability distribution of the corresponding 2D projection. This \textit{painted} point cloud, each point of dimension $(3 + C)$, is passed into the initial VoteNet. We then obtain 3D proposals and continue as before.

Compared to applying PointPainting to 3D-only approaches, PointPainting is especially effective on our framework for three reasons. First, improving the initial proposal generation stage directly raises the quality of 3D detections even after refinement - this is the usual benefit PointPainting has on 3D-only methods. Second, in our pipeline, better initial proposals mean better 2D segmentation predictions, which in turn further improves the second stage 3D refinement. Finally, we can recursively apply our entire pipeline as shown in Figure \ref{fig:recurse}. Starting with 2D segmentations from any 2D baseline, we fuse them into our first stage 3D proposal generation stage following PointPainting. Then, using these proposals, our fusion 2D network outputs even better 2D predictions, which can be used to refine these 3D proposals. In the next iteration, we then take our improved 2D predictions and fuse them into the first stage 3D proposal generation stage (previously, we had used a weaker 2D-only baseline). We continue this process and ensemble the refined 3D boxes from every iteration. We find that recursively applying our pipeline results in better 3D box predictions.

\subsection{Training Losses\label{3.6}} 

\noindent\textbf{3D Proposal Generation.} For the first stage 3D network, we simply inherit the training losses from VoteNet \cite{Qi2019DeepHV}, which can be summarized as:
\begin{equation}
    \begin{split}
        \mathcal{L}_{rpn} = &\mathcal{L}_{vote\mh reg} + \lambda_{obj\mh cls}\mathcal{L}_{obj\mh cls} + \lambda_{box}\mathcal{L}_{box} \\
        &+ \lambda_{sem\mh cls}\mathcal{L}_{sem\mh cls}
    \end{split}
     \tag{2}
\end{equation}
The loss terms correspond to the voting loss, objectness classification loss, box estimation loss, and multi-class semantic classification loss. We use the same loss weights $\lambda$ as VoteNet.

\noindent\textbf{2D Semantic Segmentation.} Our 2D segmentation model has two prediction heads - the DeepLabV3+ head and the auxiliary head. Given the multi-modality input 2D feature map $\textbf{F}^{2D\mh input}$, let $\textbf{DL3}$ and $\textbf{Aux}$ be the per-pixel segmentation predictions of the two heads and let $\textbf{Y}$ be the ground truth 2D segmentation map. Then, our 2D segmentation loss is:

\[
\mathcal{L}_{2d\mh seg} = \mathcal{L}_{ce}(\textbf{DL3}, \textbf{Y}) + \lambda_{aux}\mathcal{L}_{ce}(\textbf{Aux}, \textbf{Y}) \tag{3}
\]
where $\mathcal{L}_{ce}$ is the multi-class cross entropy loss and $\lambda_{aux}$ is set to be 0.4.

\noindent\textbf{3D Box Refinement.} The second stage 3D refinement loss consists of three parts - the box refinement loss, the multi-class semantic classification loss, and the IoU prediction loss:

\[
\mathcal{L}_{rcnn} = \mathcal{L}_{box\mh refine} + \mathcal{L}_{sem\mh cls} + \mathcal{L}_{iou} \tag{4}
\]
The box refinement loss is as follows:

\[
\mathcal{L}_{box\mh refine} = \sum_{r \in \{x, y, z, l, h, w, \theta\}}\mathcal{L}_{smooth \mh L1}(\widehat{\Delta{r}}, \Delta{r}) \tag{5}
\]
where $\Delta{r}$ is the residual between the 3D proposal box and the matched ground truth box for box parameter $r$, and $\widehat{\Delta{r}}$ is the prediction for this residual. $\mathcal{L}_{sem\mh cls}$ is multi-class cross entropy loss between the predicted class of the proposal and the ground truth box. Finally, $\mathcal{L}_{iou}$ is the confidence loss used to better predict the quality of the bounding box. The network targets the normalized 3D IoU $y$ between each proposal and its matched ground truth:

\[
y = \min(1, \max(0, 2\text{IoU} - 0.3)) \tag{6}
\]
and is trained via binary cross entropy loss.

\noindent\textbf{Overall Loss. } The total loss term is then:

\[
\mathcal{L} = \mathcal{L}_{rpn} + \mathcal{L}_{2d\mh seg} + \mathcal{L}_{rcnn} \tag{7}
\]
\label{sec:method}
\section{Experiments}
\begin{table*}[t]

\begin{center}
\resizebox{\textwidth}{!}{
\begin{tabular}{|c|c||c|c|c|c|c}
\hline
Methods & Input & AP@0.25 & AP@0.50 & AP@0.75 & mAP (AP$_{.25:.95}$) \\
\hline\hline
F-PointNet \cite{Qi2018FrustumPF} & point+RGB & 54.0 & - & - & - \\
VoteNet \cite{Qi2019DeepHV} & point & 57.7 & 32.9 & - & - \\
VoteNet \cite{Qi2019DeepHV}$^*$ & point & 58.7 & 35.1 & 1.5 & 23.8 \\
ImVoteNet \cite{Qi2020ImVoteNetB3}$^*$ & point+RGB & 64.1 & 38.7 & 2.1 & 25.8 \\
H3DNet \cite{Zhang2020H3DNet3O}$^\dagger$ & point & 61.1 & 39.0 & 3.5 & - \\
EPNet \cite{Huang2020EPNetEP} & point+RGB & 59.8 & - & - & - \\
BRNet \cite{Cheng2021BacktracingRP} & point & 61.1 & 43.7 & 5.3 & - \\
SparsePoint \cite{Liu2021SparsePointFE} & point & 61.5 & 44.2 & - & - \\
Group-Free \cite{Liu2021GroupFree3O} & point & 63.0 (62.6) & 45.2 (44.4) & - & - \\
\hline
Ours {\scriptsize (3D$\rightarrow$3D)} & point & 60.2 (59.5) & 46.0 (45.5) & 6.4 (6.5) & 29.8 (29.4) \\
Ours {\scriptsize (3D$\rightarrow$2D$\rightarrow$3D)} & point+RGB & 64.6 (64.1) & \textbf{49.0} (48.0) & 7.7 (7.9) & 31.8 (31.5) \\
Ours {\scriptsize (2D$\rightarrow$3D$\rightarrow$2D$\rightarrow$3D)} & point+RGB & 65.0 (64.5) & 48.4 (48.0) & 8.2 (7.9) & 32.0 (31.6) \\
Ours {\scriptsize (2D$\rightarrow$3D) $\times 3$} & point+RGB & \textbf{65.3} (64.7) & 48.6 (48.2) & \textbf{8.4} (8.1) & \textbf{32.2} (31.8) \\
\hline
\end{tabular}
}
\end{center}
\caption{
\textbf{Performance comparison on SUN RGB-D with state-of-the-art methods.} $^*$We report 5-times evaluation results on the checkpoint from MMdetection3D\cite{mmdet3d2020} which has higher results than the official paper. $^\dagger$ H3DNet uses 4 PointNet++ backbones.
}
\label{table:sota}
\end{table*}

\subsection{Experimental Setup\label{4.0}}
\textbf{Dataset. } We evaluate on the SUN RGB-D benchmark \cite{Song2015SUNRA, Xiao2013SUN3DAD, Janoch2011AC3, Silberman2012IndoorSA}, which is a single-view, indoor dataset for scene understanding\footnote{We focus on the SUN RGB-D dataset like most image \& point cloud fusion methods \cite{Qi2020ImVoteNetB3, Huang2020EPNetEP} instead of the similar ScanNet \cite{Dai2017ScanNetR3} dataset as ScanNet point clouds are constructed from \textit{many} RGB-D images, requiring extra processing to reconcile multiple point-to-image location correspondences.}. The dataset consists of 10,335 RGB-D images, of which 5,285 images are used for training and 5,050 are used for testing. We train and report results on the 10 most prevalent object classes following prior work \cite{Qi2020ImVoteNetB3, Zhang2020H3DNet3O}. Point clouds are generated from 2D depth maps using camera intrinsics.

\noindent\textbf{Network Architecture.}
We use the DeepLabV3+ model with an ImageNet pre-trained ResNet18 backbone for both 2D segmentation networks. Our 3D box refinement network has three linear layers of size [128, 128, 1024] before max pooling, and two shared layers $[512, 512]$ after pooling. Then, each of the 3D heads have a separate linear layer of size 512.

\noindent\textbf{Training and Inference Details.}
Our core 3D$\rightarrow$2D$\rightarrow$3D framework is trained end-to-end for 240 epochs with the ADAM optimizer with a batch size of 8. The initial learning rate is 5e-4 and is decayed 10x at 160 and 210 epochs. We follow the data augmentation strategy in \cite{Qi2020ImVoteNetB3}, sampling 20k points per view. More training details are provided in the supplementary.

\noindent\textbf{Evaluation Protocol.}
For 3D detection, Average Precision (AP) over 10 classes is reported at the standard 0.25, 0.50, and 0.75 3D IoU thresholds. However, we find that AP at specific IoU thresholds fluctuate between evaluation runs. So, we adapt the robust challenge metric used in COCO object detection \cite{Lin2014MicrosoftCC} and average the AP at 3D IoU thresholds from 0.25 to 0.95 with step size 0.05 and denote this as mAP. We prefer this metric as it fairly balances many IoU thresholds and is more stable. For 2D segmentation, we report the mIoU based on our generated 2D ground truth, ignoring overlapped regions and pixels without depth value.

\subsection{Comparison with State-of-the-art Methods\label{4.1}}
\begin{figure*}[t]
  \centering
  \includegraphics[width=\linewidth]{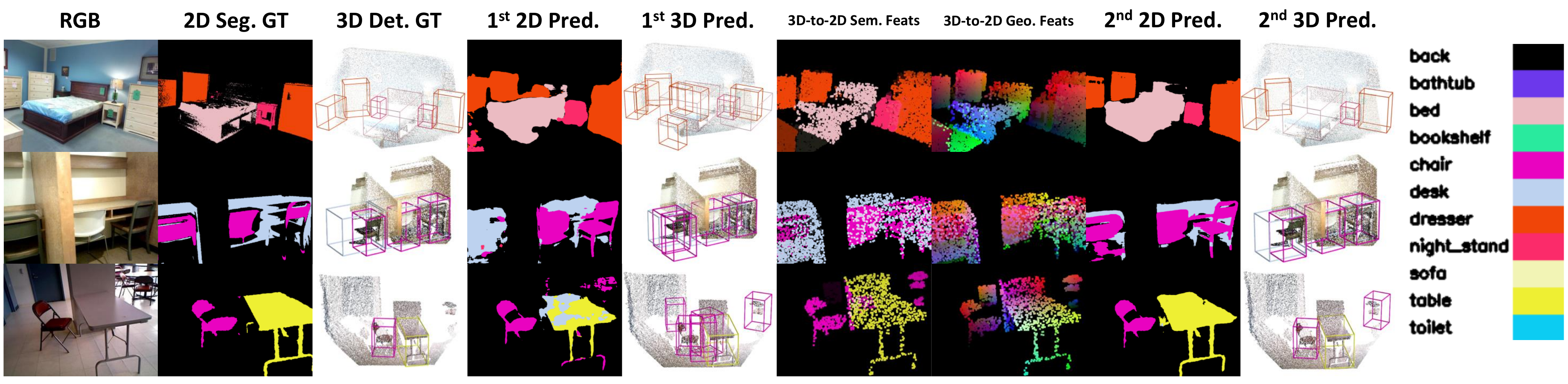}
  \caption{Qualitative results showing our alternating task modules. }
  \label{fig:vis}
\end{figure*}
\begin{table*}[t]

\begin{center}
\resizebox{.8\linewidth}{!}{
\begin{tabular}{|c|c||c|c|c|}
\hline
(Train) 2D$\rightarrow$3D & (Test) 2D$\rightarrow$3D & (Train) 2D mIoU & (Test) 2D mIoU & (Test) 3D mAP\\
\hline\hline
- & - & - & - & 29.31\\
DL3 Features & DL3 Features & - & - & 27.21\\
DL3 Seg. Preds & DL3 Seg. Preds & 89.80 & 49.36 & 29.73\\
Aux Seg. Preds & DL3 Seg. Preds & 65.37 & \textbf{50.36} & \textbf{30.00}\\
\hline
DL3 Seg. Preds & DL3 + Aux Seg. Preds & 89.80 & 50.33 & 30.81\\
Aux Seg. Preds & DL3 + Aux Seg. Preds & 65.37 & \textbf{50.84} & \textbf{31.09}\\
\hline
\end{tabular}

}
\end{center}
\vspace{.3em}
\caption{
Ablation on different methods of 2D-to-3D fusion. On the left, (Train) 2D$\rightarrow$3D denotes the fusion method during training, and (Test) the method for inference. On the right, (Train) denotes performance on train set, while (Test) denotes performance on test set.
}
\label{table:aux_2d}
\end{table*}
In this section, we compare with state-of-the-art methods. Previous works \cite{Qi2019DeepHV, Cheng2021BacktracingRP} usually train multiple times on different seeds and report the best results on the testing set. For fair comparison, we follow \cite{Liu2021GroupFree3O} in \textit{training every setting 5 times} and \textit{evaluating each setting 5 times}. The average performance over the 25 evaluations is in parentheses, and the best average evaluation result of the 5 training runs is presented on its left as the main comparison.

\noindent\textbf{Quantitative Results. } We show results in Table \ref{table:sota}. Simply by adding the 3D refinement module to VoteNet, we have a very strong 3D-only baseline, out-performing all previous methods on the AP@0.50 metric. Adding the fusion 2D segmentation model between the two 3D modules further boosts performance beyond our 3D-only baseline, improving AP@0.25 by 4.4 points, AP@0.50 by 3.0 points, and mAP by 2 points. This 3D$\rightarrow$2D$\rightarrow$3D model out-performs previous methods on all metrics. Adding another 2D segmentation model before the first 3D model improves the mAP by a small but significant margin, and recursively applying our pipeline once more as in Sec. \ref{3.5} further improves performance.

\noindent\textbf{Qualitative Results.}
In Fig. \ref{fig:vis}, we show step-by-step task predictions of our 2D$\rightarrow$3D$\rightarrow$
2D$\rightarrow$3D pipeline. We observe that predictions from the initial 2D-only model (\nth{1} 2D Pred.) are incomplete and messy. For example, in the middle row, the 2D model almost completely misses the left chair, mixing it into the desk. This leads to poor 3D boxes in the first stage 3D predictions (\nth{1} 3D Pred.) - that same chair has two boxes associated with it. However, after getting more information from the 3D detections (via 3D-to-2D features), the \nth{2} 2D predictions correctly segment the chair. These better 2D segmentations in turn lead to better 3D localization - we see that the same chair now only has a single, high-quality detection (\nth{2} 3D Pred.). Through our visualizations, we confirm that MTC-RCNN effectively leverages the mutually beneficial relationship between the 2D image and the 3D point cloud.

\subsection{Ablation Studies and Discussion\label{4.2}}
In this section, we present ablation studies to justify our training protocol and design choices. We use the stable mAP metric, averaged over the 25 evaluation runs as explained in Sec. \ref{4.1}.

\noindent\textbf{2D-to-3D: Different Methods of Fusion.}
We ablate different methods of 2D-to-3D fusion in Table \ref{table:aux_2d}. Simply fusing the 128-dim features from the DeepLabV3+ head (DL3) without 2D supervision (row 2) decreases performance compared to the two-stage 3D-only baseline (row 1) - likely due to overfitting of the 2D network. Regularizing this 2D-to-3D fusion by adding 2D supervision and fusing segmentation predictions instead (row 3) is able to out-perform the 3D-only baseline. However, there is a huge gap in mIoU between the train and test sets, causing the 3D refinement to perform poorly with imperfect test-time 2D segmentations. In the fourth row, we find that instead training on the weaker auxiliary predictions can remedy this issue, pulling train \& test mIoU closer and further boosting mAP. In the final two rows, we show that ensembling the predictions of the two heads during inference can further boost performance and that training with the auxiliary head still has better mAP.

\noindent\textbf{Two-stage 3D-only Baseline.} 
\begin{table}
\centering
\resizebox{\linewidth}{!}{
\begin{tabular}{|c|c|c||c|}
    \hline
    Method & Longer Training & 2nd Stage 3D & 3D mAP \\
    \hline\hline
    VoteNet & & & 23.81 \\
    VoteNet & \checkmark & & 24.55\\
    Ours {\scriptsize (3D$\rightarrow$3D)} & \checkmark & \checkmark & \textbf{29.31}\\
    \hline
\end{tabular}
}
\vspace{.3em}
\caption{Effects of longer training and adding 2nd stage 3D module.}
\label{table:3d}
\end{table}
In Table \ref{table:3d}, we analyze our 3D-only baseline. We first find that training VoteNet for 240 epochs instead of the 180 epochs in the original paper can improve performance. Then, adding the 2nd stage 3D refinement module significantly boosts mAP.

\noindent\textbf{Number of Points per Proposal.}
\begin{table}
\centering
\resizebox{\linewidth}{!}{
\begin{tabular}{|c|c||c|c|}
    \hline
    Method & \# Points per RoI & 2D mIoU & 3D mAP \\
    \hline\hline
    Ours {\scriptsize (3D$\rightarrow$3D)} & 512 & - & 29.31 \\
    Ours {\scriptsize (3D$\rightarrow$3D)} & 1024 & - & \textbf{29.46} \\
    Ours {\scriptsize (3D$\rightarrow$3D)} & 2048 & - & 29.43 \\
    Ours {\scriptsize (3D$\rightarrow$3D)} & 4096 & - & 29.33 \\
    \hline
    Ours {\scriptsize (3D$\rightarrow$2D$\rightarrow$3D)} & 512 & 50.84 & 31.09 \\
    Ours {\scriptsize (3D$\rightarrow$2D$\rightarrow$3D)} & 1024 & \textbf{51.35} & 31.35 \\
    Ours {\scriptsize (3D$\rightarrow$2D$\rightarrow$3D)} & 2048 & 51.03 & \textbf{31.46} \\
    Ours {\scriptsize (3D$\rightarrow$2D$\rightarrow$3D)} & 4096 & 50.06 & 31.43 \\
    \hline
\end{tabular}
}
\vspace{.3em}
\caption{Ablation on the number of points sampled in each proposal during inference.}
\label{table:num_points}
\end{table}
 In Table \ref{table:num_points}, we ablate the number of points sampled within each proposal during inference. (512 points are sampled during training). We find that the boost in 3D mAP is larger for 3D$\rightarrow$2D$\rightarrow$3D than the 3D-only model, likely due to a corresponding increase in 2D performance. We do notice, however, that from 1024 to 2048, 2D mIoU drops while 3D mAP increases, suggesting that despite the positive relationship between 2D mIoU and 3D mAP, the two many not be perfectly correlated. 

\noindent\textbf{3D-to-2D: Fusing 3D Proposals for 2D Segmentation.} 
\begin{table}
\centering
\resizebox{.8\linewidth}{!}{
\begin{tabular}{|c||c|c|}
    \hline
    Fuse 3D-to-2D & 2D mIoU & 3D mAP \\
    \hline\hline
     & 46.05 & 31.05\\
    \checkmark & \textbf{51.03} &\textbf{31.46}\\
    \hline
\end{tabular}
}
\vspace{.3em}
\caption{Fusion of 3D proposals into 2D segmentation network.}
\label{table:3d_to_2d}
\end{table}
Table \ref{table:3d_to_2d} shows that the 3D-to-2D fusion significantly improves both 2D mIoU and 3D mAP. This ablation verifies our intuition that 3D predictions can benefit 2D predictions, which can further improve 3D predictions.

\noindent\textbf{Additional Early 2D-to-3D Fusion: PointPainting} 

\begin{table}
\centering
\resizebox{\linewidth}{!}{
\begin{tabular}{|c||c|c|}
    \hline
    Method  & 2D mIoU & 3D mAP \\
    \hline\hline
    Ours {\scriptsize (3D$\rightarrow$2D$\rightarrow$3D)} & 51.03 & 31.46 \\
    Ours {\scriptsize (2D$\rightarrow$3D$\rightarrow$2D$\rightarrow$3D)} & 52.65 & 31.62 \\
    Ours {\scriptsize (2D$\rightarrow$3D$\rightarrow$2D$\rightarrow$3D} & \multirow{2}{*}{\textbf{52.93}} & \multirow{2}{*}{31.79} \\
    {\scriptsize $\rightarrow$2D$\rightarrow$3D)} &  &  \\
    Ours {\scriptsize (2D$\rightarrow$3D$\rightarrow$2D$\rightarrow$3D} & \multirow{2}{*}{52.91} & \multirow{2}{*}{\textbf{31.80}} \\
    {\scriptsize ~~~~~~~~~~~~~~~~~~$\rightarrow$2D$\rightarrow$3D$\rightarrow$2D$\rightarrow$3D)} &  &  \\
    \hline
\end{tabular}
}
\vspace{.3em}
\caption{Effects of incorporating PointPainting into our framework.}
\label{table:paint}
\end{table}
In Table \ref{table:paint}, we see that adding a 2D network (2D-only ResNet18+DeepLabV3+ achieves 46.37 mIoU) before the initial 3D proposal generation boosts both 2D mIoU and 3D mAP. Then, recursively re-using the improved 2D segmentation to generate new proposals further improves metrics. Experiments with a larger initial 2D network are in the supplementary.

\label{sec:exp}
\section{Conclusion}
In this work, we have presented a new framework that recursively uses 3D detections and 2D segmentation predictions to improve each other in a cascade fashion. Fusing semantic and geometric features extracted from 3D box proposals into a 2D segmentation network, our model generates greatly improved 2D segmentation predictions that can then be used to refine the 3D proposals. Further, by training the initial 3D proposal generator to also take as input 2D segmentation results, the entire pipeline can be repeated recursively. Our experiments demonstrate that 2D images and 3D point clouds are \textit{mutually} complementary modalities. Our network, MTC-RCNN, achieves new state-of-the-art 3D detection performance on the SUN RGB-D dataset without any 2D annotations.

\label{sec:conc}

{\small
\bibliographystyle{ieee_fullname}
\bibliography{egbib}
}

\clearpage
\appendix
\section*{Supplementary}
\section{Overview}
In this supplementary, we provide more details and results that are organized as follows:
\begin{itemize}
    \item Section \ref{supp:2dgt} provides more details and visualizations on our 2D segmentation ground truth generation.
    \item Section \ref{supp:train} explains our implementation and training details in more depth.
    \item Section \ref{supp:larger2d} presents more experiments building off of the 2D$\rightarrow$3D$\rightarrow$2D$\rightarrow$3D framework, but with a stronger initial 2D segmentation network.
    \item Section \ref{supp:morequant} contains per-category results for 3D detection.
    \item Section \ref{supp:morequalt} contains additional visualizations.
\end{itemize}

\section{2D Ground Truth Generation\label{supp:2dgt}}

\begin{figure*}[t]
  \centering
  \includegraphics[width=.60\linewidth]{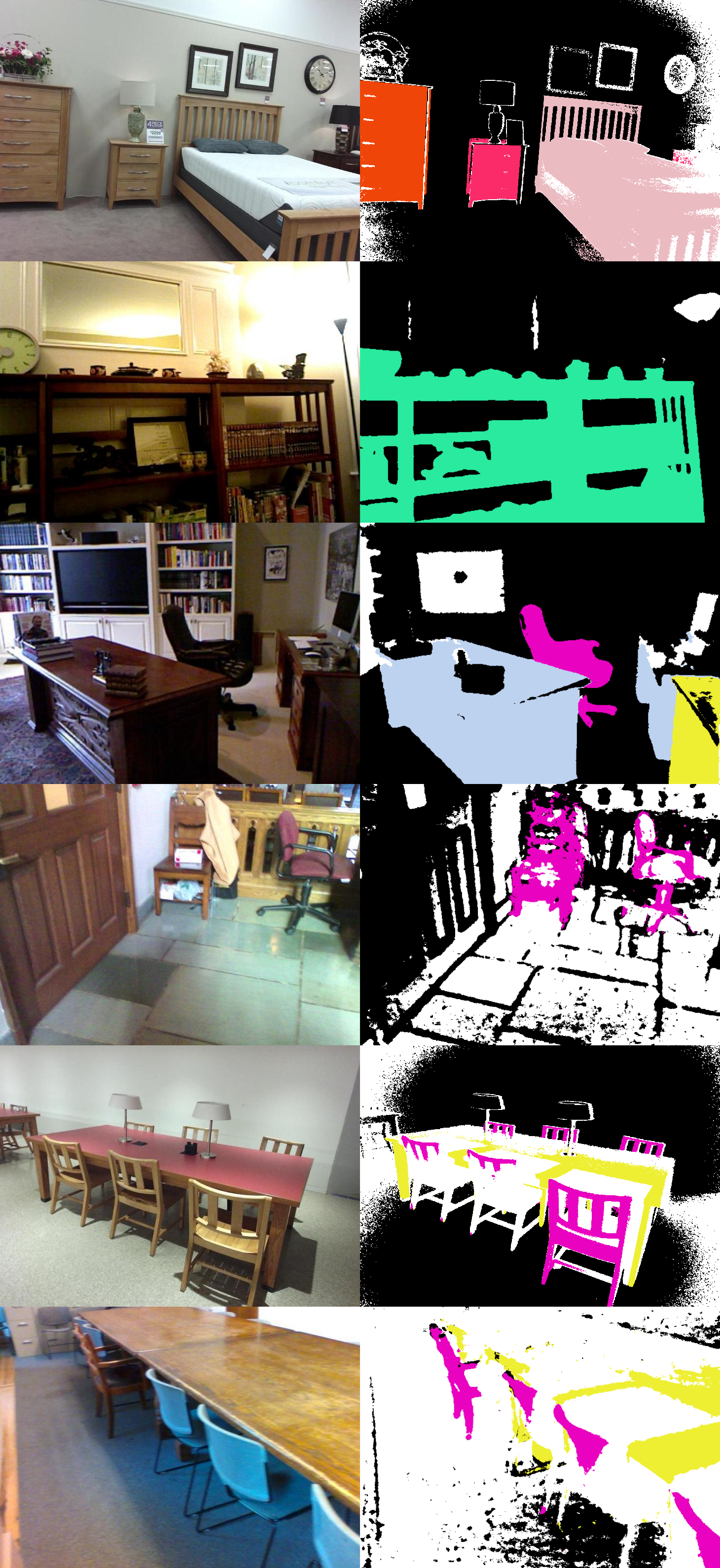}
  \caption{Visualization of generated 2D segmentation ground truth.}
  \label{fig:projgt}
\end{figure*}
To generate 2D semantic segmentation ground truth to train our 2D modules, we elect to generate them from the 3D object detection labels instead of using the 2D segmentation labels provided with the SUN RGB-D dataset. This choice enables our method to impose no additional annotation burdens and allows MTC-RCNN to be used in scenarios where 2D segmentation labels are unavailable.

To obtain the 2D labels, we expand the 2D depth map into a 3D point cloud using the camera intrinsics. Then, using the 3D box labels, if a point is within a single 3D box, the point (and its corresponding 2D pixel) is marked as the class of the 3D box it is in. If a point is not within any box, it is labeled as background. We note that some 3D points are in multiple boxes and that some 2D pixels do not have a corresponding depth value. For these latter two cases, the 2D pixel is marked to be ignored during training.

We present some example visualizations of the 2D labels in Figure \ref{fig:projgt}. Background is shown in black and ignored pixels are shown in white. We see that for some scenes (the first three rows), the segmentation labels are quite accurate, with objects clearly labeled. However, in the last three rows, we see that the generated ground truth is not always perfect. In row 4, the floor is very reflective, causing many portions of the floor to not have corresponding depth values. In row 5, the chair and table 3D bounding boxes are greatly overlapped, causing many areas to be ignored during training. In the final row we see both problems together - much of the scene is marked white. Despite these incomplete 2D segmentation labels, we find that this is enough to well-supervise MTC-RCNN, which is able to generate high-quality 2D segmentation predictions to further benefit 3D object detection.

\section{Implementation Details\label{supp:train}}
\noindent\textbf{Data Setup and Augmentation.} Following the commonly used data processing strategy from \cite{Qi2019DeepHV}, 20,000 points from each point cloud is sampled to be used as input to the network. Each point has XYZ coordinates as well as a height value (distance from the ground). The ground height is estimated from the 1\% lowest percentile of heights of points. During training, each point cloud is augmented as follows: random flip, random rotation between [-30, 30] degrees, and random scaling between [.85, 1.15]. For the 2D image, the augmentations are as follows: random resizing of image with smaller side between [480, 600], photometric distortion, ImageNet normalization of RGB channels, and depth channel normalized to between 0 and 1. Consistent with previous works, no test-time augmentation is done. To account for the randomness in sampling 20,000 points, each model is evaluated 5 times on different seeds, with the results averaged.

\noindent\textbf{Training Details.} For training, We use the AdamW optimizer ($\beta_1$=0.9, $\beta_2=0.999$) with 240 epochs. The initial learning rate is 5e-4 with a batch size of 8, and is decayed 10x at 160 and 210 epochs. Weight decay is set as 0.01 for 3D components and 1e-4 for 2D components. Gradient normalized clipping is used, with maximum norm of 10. 

We use dropout \cite{Srivastava2014DropoutAS} at multiple places in our pipeline. Dropout of rate 0.1 is used before the final classification heads in the 2D model as well as before each box parameter prediction head in the second stage 3D model. Further, dropout of rate 0.5 is used when fusing 2D segmentation predictions into 3D modules, at both 2D$\rightarrow$3D junctures in the full 2D$\rightarrow$3D$\rightarrow$2D$\rightarrow$3D pipeline. Notably, instead of randomly dropping out individual scalars, we randomly dropout 2D segmentation predictions for entire samples. More specifically, for each batch, the 2D segmentation predictions for half of the samples are not fused into 3D. We find that including this dropout allows 3D modules to not over-rely on 2D segmentation predictions, which allow them to be more robust to mistakes in 2D segmentation.

\section{2D$\rightarrow$3D$\rightarrow$2D$\rightarrow$3D with Larger Initial 2D Backbone \label{supp:larger2d}}
\begin{table*}[t]
\centering
\resizebox{.7\textwidth}{!}{
    \begin{tabular}{|c|c||c|c|}
        \hline
        First 2D Backbone & Method  & 2D mIoU & 3D mAP \\
        \hline\hline
        - & Ours {\scriptsize (3D$\rightarrow$2D$\rightarrow$3D)} & 51.03 & 31.46 \\
        \hline
        \multirow{4}{*}{ResNet18} & Initial 2D-only predictions & 46.37 & - \\
        & Ours {\scriptsize (2D$\rightarrow$3D$\rightarrow$2D$\rightarrow$3D)} & 52.65 & 31.62 \\
        & Ours {\scriptsize (2D$\rightarrow$3D$\rightarrow$2D$\rightarrow$3D$\rightarrow$2D$\rightarrow$3D)} & \textbf{52.93} & 31.79 \\\
        & Ours {\scriptsize (2D$\rightarrow$3D$\rightarrow$2D$\rightarrow$3D$\rightarrow$2D$\rightarrow$3D$\rightarrow$2D$\rightarrow$3D)} & 52.91 & \textbf{31.80} \\
        \hline
        \multirow{4}{*}{ResNet50} & Initial 2D-only predictions & 50.22 & - \\
        & Ours {\scriptsize (2D$\rightarrow$3D$\rightarrow$2D$\rightarrow$3D)} & 53.38 & 32.05 \\
        & Ours {\scriptsize (2D$\rightarrow$3D$\rightarrow$2D$\rightarrow$3D$\rightarrow$2D$\rightarrow$3D)} & \textbf{53.49} & \textbf{32.23} \\
        & Ours {\scriptsize (2D$\rightarrow$3D$\rightarrow$2D$\rightarrow$3D$\rightarrow$2D$\rightarrow$3D$\rightarrow$2D$\rightarrow$3D)} & 53.34 & 32.19 \\
        \hline
        \multirow{4}{*}{ResNet101} & Initial 2D-only predictions & 52.92 & - \\
        & Ours {\scriptsize (2D$\rightarrow$3D$\rightarrow$2D$\rightarrow$3D)} & \textbf{53.60} & 32.39 \\
        & Ours {\scriptsize (2D$\rightarrow$3D$\rightarrow$2D$\rightarrow$3D$\rightarrow$2D$\rightarrow$3D)} & 53.53 & \textbf{32.45} \\\
        & Ours {\scriptsize (2D$\rightarrow$3D$\rightarrow$2D$\rightarrow$3D$\rightarrow$2D$\rightarrow$3D$\rightarrow$2D$\rightarrow$3D)} & 53.38 & 32.39 \\
        \hline
    \end{tabular}
}
\vspace{1em}
\caption{Effects of incorporating PointPainting into our framework with larger initial 2D backbones during inference.}
\label{table:paint_large}
\end{table*}
In Table \ref{table:paint_large}, we experiment with different backbones for the initial 2D network (that is fused into the 3D proposal generation stage) during inference. During training, 2D predictions from a baseline 2D-only ResNet18 DeepLabV3+ is used. We observe that our pipeline is able to improve upon 2D predictions of even a very large backbone like ResNet101 - the initial 2D predictions achieve 52.92 2D mIoU, and the second 2D module is able to improve it significantly to 53.60, despite the second 2D module having a much smaller ResNet18 backbone. Further, we also see improvements in both 2D mIoU and 3D mAP with larger initial 2D backbone. We also experiment with recursively applying our pipeline one and two additional times (3rd and 4th row in each section) and note that although the first recursive application often yields benefits, the gains saturate or decline with the second additional application.

\section{More Quantitative Results\label{supp:morequant}}
\begin{table*}[t]

\begin{center}
\resizebox{\textwidth}{!}{
\begin{tabular}{|c|c|cccccccccc|c|}
\hline
Methods & Input & Bathtub & Bed & Bookshelf & Chair & Desk & Dresser & Nightstand & Sofa & Table & Toilet & AP@0.25 \\
\hline\hline
F-PointNet \cite{Qi2018FrustumPF} & point+RGB & 43.3 & 81.1 & 33.3 & 64.2 & 24.7 & 32.0 & 58.1 & 61.1 & 51.1 & 90.9 & 54.0 \\
VoteNet \cite{Qi2019DeepHV} & point & 74.4 & 73.0 & 28.8 & 75.3 & 22.0 & 29.8 & 62.2 & 64.0 & 47.3 & 90.1 & 57.7 \\
VoteNet \cite{Qi2019DeepHV}$^*$ & point & 74.1 & 85.8 & 34.3 & 75.6 & 26.0 & 28.3 & 60.6 & 66.7 & 50.1 & 89.6 & 58.7 \\
ImVoteNet \cite{Qi2020ImVoteNetB3} & point+RGB & 75.9 & 87.6 & 41.3 & 76.7 & 28.7 & 41.4 & \textbf{69.9} & \textbf{70.7} & 51.1 & 90.5 & 63.4 \\
H3DNet \cite{Zhang2020H3DNet3O}$^\dagger$ & point & 73.8 & 85.6 & 31.0 & 76.7 & 29.6 & 33.4 & 65.5 & 66.5 & 50.8 & 88.2 & 60.1 \\
BRNet \cite{Cheng2021BacktracingRP} & point & 76.2 & 86.9 & 29.7 & 77.4 & 29.6 & 35.9 & 65.9 & 66.4 & 51.8 & 91.3 & 61.1 \\
Group-Free \cite{Liu2021GroupFree3O} & point & \textbf{80.0} & \textbf{87.8} & 32.5 & 79.4 & \textbf{32.6} & 37.0 & 66.7 & 70.0 & 53.8 & 91.1 & 63.0 \\
\hline
Ours {\scriptsize (3D$\rightarrow$3D)} & point & 76.7 & 83.9 & 25.8 & 77.5 & 25.5 & 31.1 & 67.0 & 69.2 & 54.9 & 90.9 & 60.2 \\
Ours {\scriptsize (3D$\rightarrow$2D$\rightarrow$3D)} & point+RGB & 79.7 & 85.9 & 43.9 & 78.3 & 28.3 & 45.2 & 69.2 & 68.4 & \textbf{55.3} & 92.2 & 64.6 \\
Ours {\scriptsize (2D$\rightarrow$3D$\rightarrow$2D$\rightarrow$3D)} & point+RGB & 77.0 & 86.2 & \textbf{47.5} & 79.5 & 29.2 & \textbf{47.5} & 68.2 & 67.6 & 54.5 & 92.6 & 65.0 \\
Ours {\scriptsize (2D$\rightarrow$3D) $\times 3$} & point+RGB & 78.8 & 86.3 & 46.3 & \textbf{79.7} & 29.7 & 47.2 & 69.5 & 68.2 & 54.4 & \textbf{92.8} & \textbf{65.3} \\
\hline
\end{tabular}
}
\end{center}
\caption{
3D object detection results on SUN RGB-D. We present per-category average precision (AP) at the \textbf{0.25} IoU threshold.
$^*$We report 5-times evaluation results on the checkpoint from MMdetection3D\cite{mmdet3d2020} which has higher results than the official paper. $^\dagger$ H3DNet uses 4 PointNet++ backbones.
}
\label{table:ap_0.25}

\end{table*}
\begin{table*}[t]

\begin{center}
\resizebox{\textwidth}{!}{
\begin{tabular}{|c|c|cccccccccc|c|}
\hline
Methods & Input & Bathtub & Bed & Bookshelf & Chair & Desk & Dresser & Nightstand & Sofa & Table & Toilet & AP@0.50 \\
\hline\hline
VoteNet \cite{Qi2019DeepHV} & point & 49.9 & 47.3 & 4.6 & 54.1 & 5.2 & 13.6 & 35.0 & 41.4 & 19.7 & 58.6 & 32.9 \\
VoteNet \cite{Qi2019DeepHV}$^*$ & point & 43.0 & 54.2 & 7.3 & 54.7 & 6.0 & 13.1 & 39.4 & 49.9 & 21.6 & 62.1 & 35.1\\
H3DNet \cite{Zhang2020H3DNet3O}$^\dagger$ & point & 47.6 & 52.9 & 8.6 & 60.1 & 8.4 & 20.6 & 45.6 & 50.4 & 27.1 & 69.1 & 39.0 \\
BRNet \cite{Cheng2021BacktracingRP} & point & 55.5 & 63.8 & 9.3 & 61.6 & 10.0 & 27.3 & 53.2 & 56.7 & 28.6 & 70.9 & 43.7 \\
SparsePoint \cite{Liu2021SparsePointFE} & point & 60.9 & 63.2 & 13.8 & 61.2 & 14.2 & 23.7 & 49.1 & 57.7 & 33.2 & 65.4 & 44.2 \\
Group-Free \cite{Liu2021GroupFree3O} & point & 64.0 & 67.1 & 12.4 & 62.6 & \textbf{14.5} & 21.9 & 49.8 & 58.2 & 29.2 & 72.2 & 45.2 \\
\hline
Ours {\scriptsize (3D$\rightarrow$3D)} & point & 63.5 & 64.8 & 8.9 & 64.3 & 10.8 & 22.7 & 56.9 & 58.6 & 32.0 & 79.7 & 46.0\\
Ours {\scriptsize (3D$\rightarrow$2D$\rightarrow$3D)} & point+RGB & \textbf{64.9} & 67.0 & 20.0 & 65.8 & 11.4 & \textbf{33.9} & \textbf{57.6} & 58.1 & 34.2 & 76.9 & \textbf{49.0} \\
Ours {\scriptsize (2D$\rightarrow$3D$\rightarrow$2D$\rightarrow$3D)} & point+RGB & 54.6 & 66.2 & \textbf{23.2} & 67.0 & \textbf{12.8} & 32.6 & 54.6 & 58.6 & 34.3 & \textbf{80.2} & 48.4 \\
Ours {\scriptsize (2D$\rightarrow$3D) $\times 3$} & point+RGB & 56.1 & \textbf{67.2} & 22.5 & \textbf{67.3} & 12.7 & 32.3 & 55.5 & \textbf{59.1} & \textbf{34.4} & 78.9 & 48.6\\
\hline
\end{tabular}
}
\end{center}
\caption{
3D object detection results on SUN RGB-D. We present per-category average precision (AP) at the \textbf{0.50} IoU threshold.
$^*$We report 5-times evaluation results on the checkpoint from MMdetection3D\cite{mmdet3d2020} which has higher results than the official paper. $^\dagger$ H3DNet uses 4 PointNet++ backbones.
}
\label{table:ap_0.50}
\end{table*}
\begin{table*}[t]

\begin{center}
\resizebox{\textwidth}{!}{
\begin{tabular}{|c|c|cccccccccc|c|}
\hline
Methods & Input & Bathtub & Bed & Bookshelf & Chair & Desk & Dresser & Nightstand & Sofa & Table & Toilet & AP@0.75 \\
\hline\hline
VoteNet \cite{Qi2019DeepHV}$^*$ & point & 0.8 & 4.0 & 0.0 & 2.5 & 0.0 & 0.1 & 0.2 & 4.3 & 0.5 & 2.6 & 1.5\\
BRNet \cite{Cheng2021BacktracingRP} & point & - & - & - & - & - & - & - & - & - & - & 5.3 \\
\hline
Ours {\scriptsize (3D$\rightarrow$3D)} & point & \textbf{5.1} & 18.2 & 0.1 & 8.1 & 0.4 & 1.4 & 2.0 & 15.3 & 2.3 & 10.8 & 6.4 \\
Ours {\scriptsize (3D$\rightarrow$2D$\rightarrow$3D)} & point+RGB & 4.1 & \textbf{25.1} & 0.4 & 8.8 & 0.5 & \textbf{3.5} & 3.7 & 16.8 & 2.5 & 12.0 & 7.7 \\
Ours {\scriptsize (2D$\rightarrow$3D$\rightarrow$2D$\rightarrow$3D)} & point+RGB & 4.0 & 22.8 & 0.3 & 9.1 & \textbf{1.1} & 2.6 & 4.5 & 17.2 & \textbf{3.8} & \textbf{16.6} & 8.2 \\
Ours {\scriptsize (2D$\rightarrow$3D) $\times 3$} & point+RGB & 4.6 & 23.0 & \textbf{0.5} & \textbf{9.2} & \textbf{1.1} & 2.9 & \textbf{4.7} & \textbf{18.2} & \textbf{3.8} & 16.4 & \textbf{8.4}\\
\hline
\end{tabular}
}
\end{center}
\caption{
3D object detection results on SUN RGB-D. We present per-category average precision (AP) at the \textbf{0.75} IoU threshold.
$^*$We report 5-times evaluation results on the checkpoint from MMdetection3D\cite{mmdet3d2020} which has higher results than the official paper.
}
\label{table:ap_0.75}
\end{table*}
\begin{table*}[t]

\begin{center}
\resizebox{\textwidth}{!}{
\begin{tabular}{|c|c|cccccccccc|c|}
\hline
Methods & Input & Bathtub & Bed & Bookshelf & Chair & Desk & Dresser & Nightstand & Sofa & Table & Toilet & mAP (AP$_{.25:.95}$) \\
\hline\hline
VoteNet \cite{Qi2019DeepHV}$^*$ & point & 27.0 & 36.7 & 9.4 & 34.2 & 6.7 & 10.0 & 24.9 & 32.0 & 17.1 & 40.3 & 23.8\\
\hline
Ours {\scriptsize (3D$\rightarrow$3D)} & point & 38.3 & 44.0 & 8.8 & 40.4 & 8.5 & 13.7 & 33.2 & 39.1 & 22.8 & 49.3 & 29.8\\
Ours {\scriptsize (3D$\rightarrow$2D$\rightarrow$3D)} & point+RGB & \textbf{40.0} & \textbf{46.4} & 14.5 & 41.1 & 9.3 & \textbf{21.4} & 34.5 & 38.7 & 23.4 & 49.1 & 31.8 \\
Ours {\scriptsize (2D$\rightarrow$3D$\rightarrow$2D$\rightarrow$3D)} & point+RGB & 36.5 & 45.7 & \textbf{16.9} & 41.8 & \textbf{10.1} & 21.1 & 33.7 & 38.8 & 23.8 & \textbf{51.1} & 32.0\\
Ours {\scriptsize (2D$\rightarrow$3D) $\times 3$} & point+RGB & 37.6 & 46.1 & 16.6 & \textbf{42.0} & \textbf{10.1} & 20.9 & \textbf{34.7} & \textbf{39.3} & \textbf{23.9} & 50.8 & \textbf{32.2}\\
\hline
\end{tabular}
}
\end{center}
\caption{
3D object detection results on SUN RGB-D. We present per-category mean average precision (mAP), averaged over APs from IoUs from 0.25 to 0.95 at 0.05 intervals.
$^*$We report 5-times evaluation results on the checkpoint from MMdetection3D\cite{mmdet3d2020} which has higher results than the official paper.
}
\label{table:mAP}
\end{table*}
In Tables \ref{table:ap_0.25}, \ref{table:ap_0.50}, \ref{table:ap_0.75}, \ref{table:mAP}, we report per-category evaluation results. The commonly used AP@0.25 metric is quite saturated by recent works, so we also report AP@0.50, AP@0.75, and mAP (AP$_{.25:.95}$). mAP is a tougher and more stable metric incorporating many IoU thresholds. We observe that our 3D-only (3D$\rightarrow$3D) model is already a very strong state-of-the-art method. Further, including additional 2D segmentation between the two 3D modules boosts performance by a significant margin. Then, also including a 2D network before the first 3D proposal generation stage further boosts performance (AP@0.25, AP@0.75, mAP). Finally, recursively applying our pipeline once more demonstrates a small but consistent additional boost in all metrics. 

\section{More Qualitative Results\label{supp:morequalt}}
\begin{figure*}[t]
  \centering
  \includegraphics[width=\linewidth]{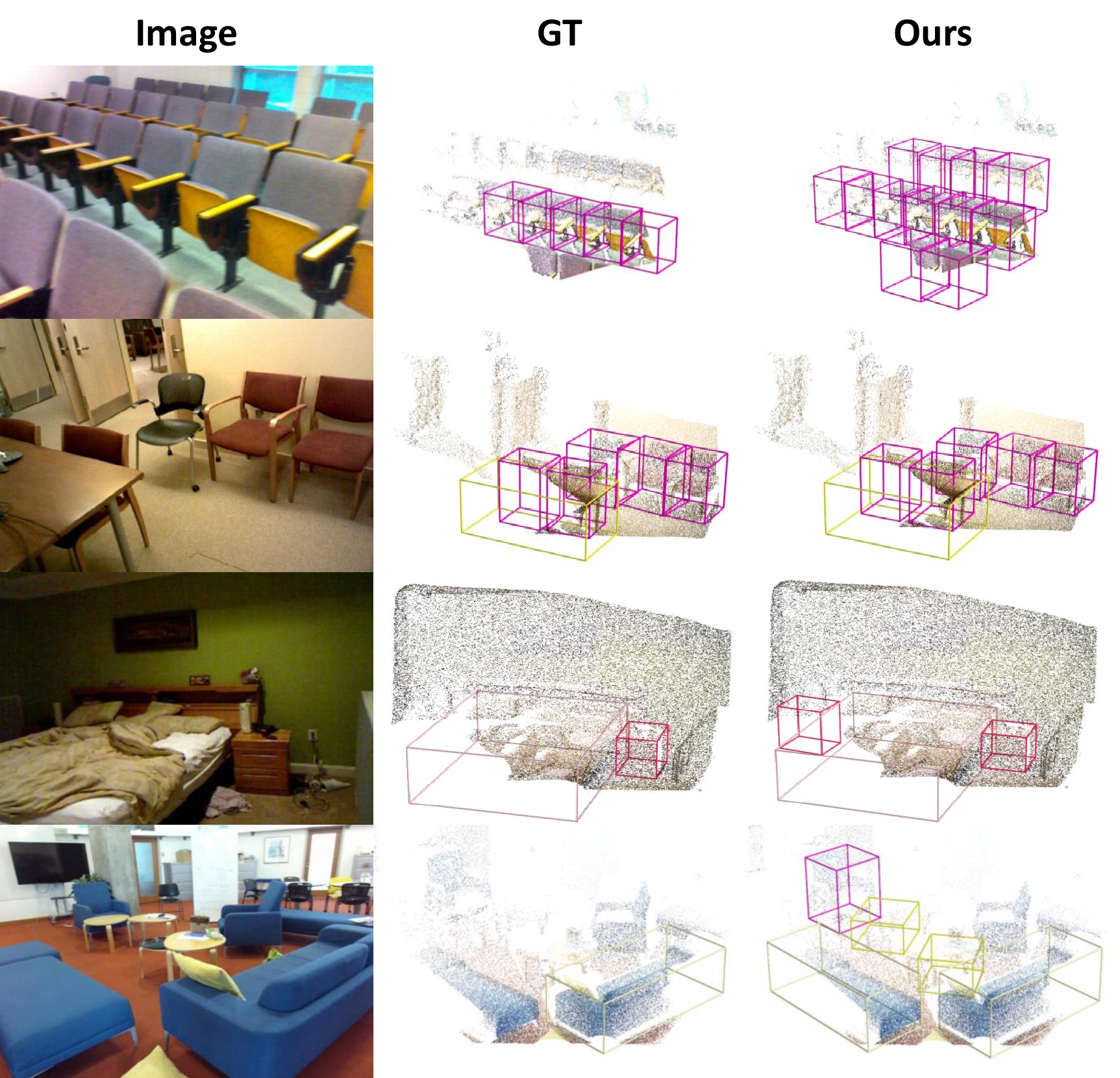}
  \caption{Additional qualitative results from our model.}
  \label{fig:add_vis}
\end{figure*}
We show additional 3D detection results from our pipeline in Figure \ref{fig:add_vis}. We find that our method is able to produce highly accurate predictions, often capturing objects not labeled in ground truth.

\end{document}